%% file: main.tex
\documentclass{article}
\usepackage[letterpaper,top=2cm,bottom=2cm,left=3cm,right=3cm,marginparwidth=1.75cm]{geometry}
\usepackage[english]{babel}

\usepackage{tikz}
\usepackage{comment}
\usepackage{amsmath,amssymb}
\usepackage{graphicx}
\usepackage{color}
\usepackage{booktabs}
\usepackage{float}
\usepackage{wrapfig}
\usepackage{cite}
\usepackage[colorlinks=true, allcolors=blue]{hyperref}

\def\mbf#1{\mathbf{#1}}
\def\mbb#1{\mathbb{#1}}

\def\mcal#1{\mathcal{#1}}
\def\bs#1{\boldsymbol{#1}}
\def\myx{\mbf{x}}

\def\myz{\mbf{z}}

\def\myw{\mbf{w}}
\def\mytheta{\bs{\theta}}
\def\myphi{\bs{\phi}}

\newcommand{\etal}{\textit{et al.}}

\usepackage{algpseudocode}
\usepackage{algorithm}

\algrenewcommand\algorithmicrequire{\textbf{Input:}}
\algrenewcommand\algorithmicensure{\textbf{Output:}}

\algnewcommand{\Inputs}[1]{%
  \State \textbf{Inputs:}
  \Statex \hspace*{\algorithmicindent}\parbox[t]{.8\linewidth}{\raggedright #1}
}
\algnewcommand{\Init}[1]{%
  \State \textbf{Initialization:}
  \Statex \hspace*{\algorithmicindent}\parbox[t]{.8\linewidth}{\raggedright #1}
}
\algnewcommand{\Outputs}[1]{%
  \State \textbf{Output:}
  \Statex \hspace*{\algorithmicindent}\parbox[t]{.8\linewidth}{\raggedright #1}
}

\usepackage{makecell}
\usepackage{xcolor}

\usepackage{authblk}
\providecommand{\keywords}[1]{\noindent\textbf{Keywords} #1}
\newcommand{\method}{HiT-DVAE}

\title{HiT-DVAE: Human Motion Generation via Hierarchical Transformer Dynamical VAE}

\author[1*]{Xiaoyu Bie}
\author[1,4*]{Wen Guo}
\author[2]{Simon Leglaive}
\author[3]{Lauren Girin}
\author[4]{Francesc Moreno-Noguer}
\author[1]{Xavier Alameda-Pineda}

\affil[1]{Inria, Univ. Grenoble Alpes, CNRS, Grenoble INP, LJK, France}
\affil[2]{CentraleSup\'elec, IETR, France}
\affil[3]{Univ. Grenoble Alpes, Grenoble-INP, GIPSA-lab, France}
\affil[4]{Institut de Robòtica i Informàtica Industrial, CSIC-UPC, Spain}
\date{}

\begin{document}

\maketitle

\renewcommand{\thefootnote}{\fnsymbol{footnote}}
\footnotetext[1]{Equal contribution.}
\footnotetext[2]{This research was supported by ANR-3IA MIAI (ANR-19-P3IA-0003), ANR-JCJC ML3RI (ANR-19-CE33-0008-01), H2020 SPRING (funded by EC under GA \#871245),  by the Spanish government with the project MoHuCo PID2020-120049RB-I00 and  by an Amazon Research Award. This work was performed using HPC resources from the ``M\'esocentre'' computing center of CentraleSup\'elec and \'Ecole Normale Sup\'erieure Paris-Saclay supported by CNRS and R\'egion \^Ile-de-France. We also thank Nvidia for hardware donation under the Academic Hardware Grant Program.}
% \footnotetext[2]{Univ. Grenoble Alpes, CNRS, Grenoble INP, LJK, 38000 Grenoble, France}
% \footnotetext[3]{CSIC-UPC, Barcelona, Spain}
\renewcommand*{\thefootnote}{\arabic{footnote}}

\begin{abstract}
Studies on the automatic processing of 3D human pose data have flourished in the recent past. In this paper, we are interested in the generation of plausible and diverse future human poses following an observed 3D pose sequence. Current methods address this problem by injecting random variables from a single latent space into a deterministic motion prediction framework, which precludes the inherent multi-modality in human motion generation. In addition, previous works rarely explore the use of attention to select which frames are to be used to inform the generation process up to our knowledge. To overcome these limitations, we propose Hierarchical Transformer Dynamical Variational Autoencoder, \method{}, which implements auto-regressive generation with transformer-like attention mechanisms. \method{} simultaneously learns the evolution of data and latent space distribution with time correlated probabilistic dependencies, thus enabling the generative model to learn a more complex and time-varying latent space as well as diverse and realistic human motions. Furthermore, the auto-regressive generation brings more flexibility on observation and prediction, i.e. one can have any length of observation and predict arbitrary large sequences of poses with a single pre-trained model. We evaluate the proposed method on HumanEva-I and Human3.6M with various evaluation methods, and outperform the state-of-the-art methods on most of the metrics.


\end{abstract}

%%%%%%%%%%%%%%%%%%%%%%%%%%%%%%%%%%%%%%%%
%% BODY TEXT

\input{1-Introduction}
\input{2-RelatedWork}
\input{3-Method}
\input{4-Experiments}

%%%%%%%%%%%%%%%%%%%%%%%%%%%%%%%%%%%%%%%%
\section{Conclusions}

In this paper we have investigated the use of attention combined with temporal probabilistic models for human motion generation. In particular, we proposed \method{}, a variational method modeling temporal dependencies between the observations and the latent variables, and exploiting attention to select which observations will be used to inform the generation of the current frame. Up to our knowledge, this is the first time that models with temporal latent variables and the use of attention are proposed to handle the human motion generation task. 
We exhaustively evaluated our method on two widely used datasets, HumanEva and Human3.6M, and reported state-of-the-art results.%, specially in terms of medium errors and implicit metrics, which demonstrate that our model could stable generate reasonable actions with good performance. \francescrmk{do not understand the last sentence}
%%%%%%%%%%%%%%%%%%%%%%%%%%%%%%%%%%%%%%%%
\bibliographystyle{splncs04}
\bibliography{egbib}
\clearpage
\input{5-Appendix}

\end{document}

%% file: 1-Introduction.tex
\section{Introduction}

Modeling 3D human pose and motion is a fundamental problem towards understanding human behaviour, with a wide range of applications in medical prognosis, 3D content production, autonomous driving and  human-robot interaction. One of the most studied  computer vision tasks is pose estimation either from single images~\cite{toshev2014deeppose,moreno20173d,pavlakos2017coarse,guo2021pi}, monocular videos~\cite{pfister2015flowing,pfister2015flowing,pavllo20193d}, or from multi-camera settings~\cite{amin2013multi,rhodin2018learning,tu2020voxelpose}. This development is certainly due to the availability of large-scale datasets such as Human3.6M~\cite{h36m_pami}. Beyond extracting 3D human pose information from various types of visual data, a number of works have been proposed around the development of machine learning methods allowing to forecast future motion~\cite{fragkiadaki2015recurrent,jain2016structural,martinez2017human,liu2019towards,chiu2019action,butepage2017deep,butepage2018anticipating,holden2015learning,hernandez2019human,li2018convolutional,gui2018adversarial,mao2019learning,kipf2016semi,elickovic2018graph,mao2020history,li2021rain,Dang_2021_ICCV,Sofianos_2021_ICCV,adeli2020socially,Adeli_2021_ICCV,guo2021multi} and more recently on generating plausible future sequences of realistic human 3D pose data~\cite{lin2018human,kundu2019bihmp,barsoum2018hp,mao2021gsps,mao2019learning,walker2017pose,yan2018mt,aliakbarian2020stochastic,yuan2020dlow,aliakbarian2021contextually,petrovich2021action,cai2021unified,guo2020action2motion}. 

\input{fig_teaser}
%In the recent past the processing of 3D human pose data has attracted lots of attention. One of the closely related computer vision tasks is the estimation of the skeleton joints either from single images~\cite{}, \francesc{monocular videos~\cite{}}, in multi-camera settings~\cite{}, or from RGB-D data~\cite{}. Such development is certainly due to the availability of large-scale datasets such as \xavirmk{BLABLABLA}. Beyond extracting 3D human pose information from various types of visual data, in the recent past much research has fostered around the study of the 3D pose data \textit{per se}. One the of prominent challenges in this regard is the development of machine learning methods allowing to generate sequences of realistic human 3D pose data.

Human motion generation has several challenges:
i) diversity: different from deterministic prediction, the generation methods should not just learn average patterns, but need to faithfully reflect the intrinsic intra-class variability. 
ii) dynamics: the generation process must inherently model the dynamics of the 3D human pose data, so that they can be transferred to the generated data and avoid collapsing to a stable motion. 
iii) smoothness: the generated data should be smooth.

Some of these challenges have been partially addressed in past studies, although up to our knowledge, there is no existing methodology designed to face all  three challenges.
For instance, MT-VAE~\cite{yan2018mt} combined the RNN-based motion prediction model with conditional variational autoencoders (VAE), where the difference  between the observed and future poses was encoded into the latent variable, which was then  concatenated with the RNN's hidden state to account for the dynamics.
DLow~\cite{yuan2020dlow} proposed to generate a diverse set of motion sequences, by training fifty different encoders (and a single decoder), and obtaining fifty different instances of the latent variable, and thus fifty different generated motions. 
GSPS~\cite{mao2021gsps} inherited the diversity loss from DLow, but utilized a more powerful motion prediction framework, based on graph convolutional network (GCN) rather than the RNN. Diversity was obtained by concatenating random noise to the observed sequences. Finally, ACTOR, a Transformer-VAE method, was proposed in~\cite{petrovich2021action} to perform  sequential generation, as opposed to previous methods that  were designed for a fixed output length. 
ACTOR~\cite{petrovich2021action} learned an action class informed token as input to the encoder and decoder, thus conditioning the generation with manually annotated labels. 
All the above methods encode the observed human poses into sequence-level embedding, meaning that the entire observed sequence is encoded into a single time-independent embedding.

% \begin{wrapfigure}{R}{0.55\textwidth}
% \begin{table}[t]
%     \label{table:comp}\vspace{-6mm}
%     \caption{Comparison of different methods dealing with different challenges. Our method is the only one deals with all the three challenges.}
%     % \label{tab:my_label}
%     \centering
%     %\vspace{-6mm}
%     \resizebox{0.55\textwidth}{!}{
%     \begin{tabular}{l|cccc|c}
%     \toprule
    
%      & DLow~\cite{yuan2020dlow} & GSPS~\cite{mao2021gsps} & ACTOR~\cite{petrovich2021action} &  MT-VAE~\cite{yan2018mt} &ours\\
%      \midrule
%     diversity & \checkmark& \checkmark& & &\checkmark\\
%     dynamics & \checkmark & \checkmark & \checkmark & \checkmark&\checkmark\\
%     smoothness &  &  & \checkmark& &\checkmark\\ 
%      \bottomrule
%     \end{tabular}
%     }\vspace{-4mm}
% \end{table}
% \end{wrapfigure}

This motivates us to propose \method{} inspired by the recent literature on dynamical variational autoencoders (DVAE)~\cite{girin2020dynamical}. 
Our method belongs to the very general family of variational autoencoders (VAE) and is therefore a probabilistic method inherently able to generate stochastic, and hence diverse output. 
More precisely, using DVAE, the sequence of observations is encoded into a sequence of latent variables instead of a single latent variable, thus offering larger representation power to learn and exploit the motion dynamics, see Figure~\ref{fig:data_space}. 
Besides, we model the generative process with auto-regressive dependencies, meaning that the generation of each frame depends on the previous ones and can be done sequentially. We can then train this model to generate sequences of arbitrary length. 
Finally, we implement these auto-regressive dependencies with a transformer-like (attention-based) encoder-decoder architecture, thus learning to automatically select which are the best frames to inform the generation of the next human 3D pose. 
Overall, \method{} implements auto-regressive probabilistic dependencies with transformer-like attention mechanisms enabling the learning of pose sequence dynamics as well as stochastic motion generation. Hence, the proposed solution deals with all the three challenges mentioned above.

%% evaluation
Considering the evaluation of generated data, recent works either evaluate the generations directly on the joint location of poses~\cite{yuan2020dlow,mao2021gsps}, or using the feature extracted from a pretrained feature extractor~\cite{petrovich2021action,guo2020action2motion}. Both protocols have clear shortcomings: the former just evaluates the best generated sample and the diversity of all generations, ignoring the performance of the generations except the best one; while the latter depends on the quality of the feature extractor. To thoroughly evaluate the generation quality, we use both evaluation methods, and broaden the first one to take performance stability into consideration.

%% result
We thoroughly evaluate \method{} on HumanEva-I and Human3.6M datasets, using both explicit and implicit metrics to measure the quality of generated data.
Experimental results show that our method achieves state-of-the-art on most of the metrics for both datasets, proving that the generation of \method{} has high quality (smaller errors, better features, correct action) with better performance stability. 

%% file: fig_teaser.tex
\begin{figure}[htbp]
    \centering
    \includegraphics[width=0.99\linewidth]{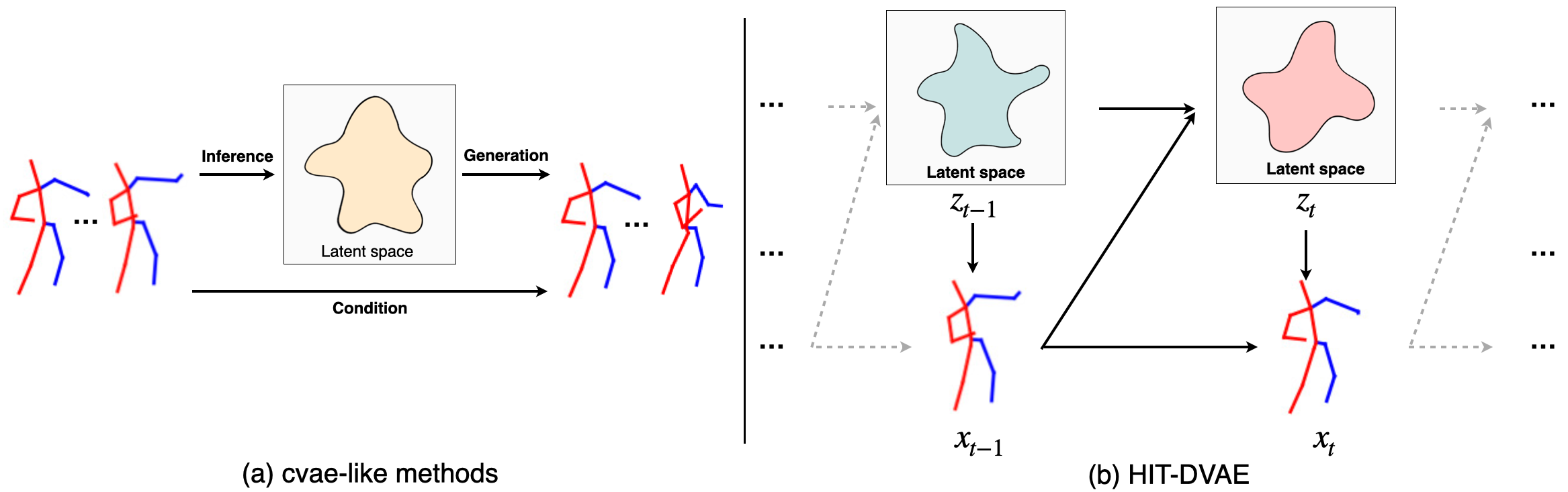}
    \caption{Conditional variational autoencoder (CVAE) based methods (left) aim to encode a sequence of observations into a single latent embedding and learn to generate future motions by combining samples from this single latent space with past observations. Our method HiT-DVAE (right) aims to learn a sequential generative model on joint distribution of data and latent variables. Diverse motion generation is conducted by alternate use of generation on $\myz_t$ and $\myx_t$.}
    \label{fig:data_space}
\end{figure}

%% file: 2-RelatedWork.tex
\section{Related Work}

%%%%%%%%%%%%%%%%%%%%%%%%%%%%%%%%%%%%%%%%
\subsection{Modeling Future Human motion}
\label{sec:realated_work}
The forecasting of human motion has been addressed under two paradigms so far: deterministic motion prediction and stochastic human motion generation. 
%There are two tasks for modeling future human motions: deterministic motion prediction and stochastic human motion generation.
The former aims at using deterministic approaches to regress a single future motion from the past observation which is the most likely to the ground truth; while the latter focuses at generating various possibilities of the future to model the multi-modal nature of human motions.\\

% \noindent\textbf{Deterministic human motion prediction.}
\subsubsection{Deterministic human motion prediction}
Due to the inherent sequential structure of human motion, 3D human motion prediction has been mostly addressed with recurrent models~\cite{fragkiadaki2015recurrent,jain2016structural,martinez2017human}. 
However, although RNNs can achieve great success in motion prediction, they represent the entire past motion history with a fixed-size hidden state and tend to converge to a static pose. Some works alleviate this problem by using RNN variants~\cite{liu2019towards,chiu2019action}, sliding windows~\cite{butepage2017deep,butepage2018anticipating}, convolutional models~\cite{holden2015learning,hernandez2019human,li2018convolutional} or adversarial training~\cite{gui2018adversarial}.
Since human body pose data are structured, directly encoding the whole body into a compact latent embedding neglects the spatial connectivity of human joints. To this end, recent work tend to leverage the forward graph convolutional network (GCN)~\cite{kipf2016semi,elickovic2018graph} with a predefined or learnable adjacency matrix~\cite{mao2019learning,mao2020history,Dang_2021_ICCV,Sofianos_2021_ICCV,li2021rain,adeli2020socially,Adeli_2021_ICCV}. While deterministic methods have achieved promising results on accurate predictions, they exhibit strong limitations when it comes to model the diversity of plausible human motion forecasts. Stochastic methods are promising tools to overcome these limitations.

% lin2018human,kundu2019bihmp,barsoum2018hp,mao2021gsps,mao2019learning,
% walker2017pose,yan2018mt,aliakbarian2020stochastic,yuan2020dlow,aliakbarian2021contextually,petrovich2021action,cai2021unified,guo2020action2motion

\subsubsection{Stochastic human motion generation}  To generate multiple future outcomes given a sequence of past observations, two types of approaches have been studied in the recent past: (i) the enhancement of deterministic methods with stochastic variations, e.g., incorporating noise, and (ii) leveraging conditional variational architectures that learn a probability distribution. In the first category, early works include combining random noise with hidden states either by concatenation~\cite{lin2018human,kundu2019bihmp} or addition~\cite{barsoum2018hp}. More recently, Mao~\etal~\cite{mao2021gsps} further investigated this paradigm with a GCN based motion prediction model~\cite{mao2019learning}, and showed promising results with dedicated designed losses. In the second category, past observations are encoded to learn a posterior latent space, then a random variable will be sampled and then combined with observations to predict the future~\cite{walker2017pose,yan2018mt,aliakbarian2020stochastic,cai2021unified,aliakbarian2021contextually}. Rencently, DLow~\cite{yuan2020dlow} proposed to explicitly generate a large number of samples during training, then to use a energy function to promote the diverse generation. ACTOR~\cite{petrovich2021action} first introduce a Transformer-VAE to obtain long term attention. Rather than modeling the whole observation into a single embedding, Action2Motion~\cite{guo2020action2motion} and HuMor~\cite{rempe2021humor} exploit a auto-regressive generative model that the current generation will depend on the past prediction. However, they do not model the entire sequence, but only on the last frame, thus resulting in non-smooth motion generation.

%%%%%%%%%%%%%%%%%%%%%%%%%%%%%%%%%%%%%%%%
\subsection{Deep generative modeling} %\biermk{a brief intro to DVAE}
The family of stochastic human motion generation methods are mostly based in the general paradigm of variational inference, and of variational autoencoders. VAEs model the joint distribution of an observation $\myx$ and a latent variable $\myz$. In stochastic human motion generation, the observations $\myx$ often corresponds to a sequence of poses, rather than a single pose. However, up to our knowledge, most of the previous methods use a single latent variable $\myz$ to encode the entire observed sequence. Alternatively, one could consider a sequence of latent variables and of observations, and use a VAE to model the relationship between between $\myx_t$ and $\myz_t$ without any time dependencies. However, the dynamics and any temporal relationships cannot be modeled in this case, which is obviously not desirable. Dynamical variational autoencoders (DVAEs)~\cite{girin2020dynamical} offer the possibility to model data sequences within the general paradigm of variational inference. DVAE is a general class of models and different models are obtained when considering various dependencies between the variables, e.g., variational recurrent neural networks~\cite{chung2015recurrent} or stochastic recurrent neural networks~\cite{fraccaro2016sequential}. However, current DVAE models have a major limitation: the probabilistic dependencies between variables are always implemented with recurrent neural networks (or variants), thus avoiding the possibility to select which past frames are used to inform the generation of the current frame. In addition and up to our knowledge, the use of DVAEs for human motion forecasting has not been investigated so far. This motivates us to explore the use of attention mechanisms within the DVAE paradigm with applications to human motion forecasting, as explained in the following.

% To bridge this gap, we propose \method{}, as explained in the following.

%% file: 3-Method.tex
\section{Method}

%%%%%%%%%%%%%%%%%%%%%%%%%%%%%%%%%%%%%%%%
% \subsection{Problem Definition and Background}
We address the problem of 3D human motion generation that we formalise as follows. 
Given a sequence of $O$ observed 3D poses of a person $\myx_{1:O} = [
\myx_1, \ldots, \myx_O]$, our aim is to generate a sequence of $G$ 3D poses $\myx_{O+1:O+G} = [\myx_{O+1}, \ldots \myx_{O+G}]$, that follow the observations $\myx_{1:O}$. Each pose vector  $\myx_t \in \mathbb{R}^{J\times3}$ encodes the location of the $J$ joints of a person at time $t$ in Cartesian coordinates. Different from deterministic human motion prediction, we intend to generate multiple plausible future motion sequences with arbitrary length. To this end, we propose a new method named Hierarchical Transformer Dynamical Variational AutoEncoder or \method{}. Our method is based on the recently reviewed family of dynamical variational autoencoders~\cite{girin2020dynamical}, which formulates the generative process of time series in an autoregressive and time-dependent perspective.
%\francescrmk{what does 'time-independent' mean in this context? arbirary length? we can start predicting at frame T+1 or T+K?}\biermk{this means the generative model has time dependencies, e.g. $x_t$ depends on $x_{1:t-1}$ and $z_{t}$, while in the vanilla VAE, $x_t$ only depends on $z_t$}. 
Up to our knowledge, this general methodology has never been combined with attention-based mechanisms. On the one hand, existing variants of DVAEs are always implemented with recurrent networks~\cite{girin2020dynamical} (or standard variants such as LSTM and GRU). On the other hand, even if self-attention has been proven useful when combined with a Conditional VAE~\cite{petrovich2021action}, the architectures proposed so far encode the entire sequence into a single latent variable $\myz$, therefore potentially limiting the representation capabilities of temporal dynamics. We propose \method{} to get the best of both worlds, enabling stochastic motion generation together with dynamic sequence modeling via transformer-like attention mechanisms.

%\subsection{\method{}: Hierarchical transformer-based dynamical variational autoencoder}

\subsection{\method{}}

% \begin{figure*}[t]
%     \centering
%     \includegraphics[width=0.9\linewidth]{TransDVAE.png}
%     \caption{TransDVAE architecture.\biermk{temporary} \xavirmk{1) There is a concatenation before the transformer encoder, right? 2) The concatenation of the transformer decoder of $\myz$ should lead to $[f(\myx_t),\myw]$ } \simonrmk{in the decoder on the left, $x_{1:T}$ is used as the keys and values and no masking process appears, so it gives the impression that $x_t$ is generated from $x_{1:T}$. Maybe indicate somewhere that a mask is applied? (similar remark for the decoder on the right)}}
%     \label{fig:HIT-DVAE}
% \end{figure*}

% \input{fig_pipeline}

The proposed method is based on the very general DVAE methodology (see~\cite{girin2020dynamical} for an exhaustive presentation on the topic). The basic principle of DVAEs is that for every observation $\myx_t$ there is a corresponding latent variable $\myz_t$, as opposed to VAEs which would encode the entire observed sequence $\myx_{1:O}$ into a single latent variable $\myz$. % \francescrmk{does this mean that if we want to generate a sequence of length K we need to observe K frames in the past?}\biermk{For CVAE, this means we need to observe O frames in the past, O and K don't need to be the same, but they need to be defined and fixed before training and inference}. 
The sequence of observations and corresponding latent variables will be denoted by $\myx_{1:T} = [\myx_t]_{t=1}^{T}$ and $\myz_{1:T} = [\myz_t]_{t=1}^{T}$, respectively. For the time being, we will assume that $T=O+G$, as if all 3D poses were observed even if this is not our setting. We will discuss the impact of having hybrid half-observed half-generated sequences later on.

In addition to the time-dependent latent variable $\myz_{1:T}$, and inspired by~\cite{petrovich2021action,yingzhen2018disentangled}, we add a time-independent latent variable $\myw$. Very differently from~\cite{petrovich2021action}, $\myw$ will be learned in an unsupervised manner within the DVAE methodology, see~\cite{yingzhen2018disentangled}, thus without requiring action class labels. Formally, the proposed generative model writes:
% \xavirmk{The dependencies are those of our method, and are not motivated. Also this is not ``a generic DVAE''.}
% be the corresponding latent variables, a DVAE is defined by the joint probability density function (pdf) of observed and latent sequences ($\myx_{1:T},\myz_{1:T}$). Follow \cite{yingzhen2018disentangled}, we further add a time-independent concept $\myw$. This global joint pdf can be factorized using the chain rule and with some simplifications:
\begin{align}
    p_{\bs{\theta}}(\myx_{1:T}, \myz_{1:T}, \myw) &= \prod_{t=1}^{T} p_{\bs{\theta}}(\myx_{t}, \myz_{t}, \myw | \myx_{1:t-1},\myz_{1:t-1}) \\
    &= p_{\bs{\theta}_{\myw}}(\myw) \prod_{t=1}^{T} p_{\bs{\theta}_{\myx}}(\myx_{t} | \myx_{1:t-1},\myz_t, \myw) p_{\bs{\theta}_{\myz}}(\myz_{t} | \myx_{t-1},\myz_{1:t-1}, \myw),
\label{eq:dvae_generation}
\end{align}
meaning that the generative processes of both the observed and latent variables are auto-regressive, with cross-dependencies. We set $\bs{\theta} = \bs{\theta}_\myw \cup \bs{\theta}_\myz \cup \bs{\theta}_\myx$. In order to learn this generative model, we introduce an inference model ($\bs{\phi}=\bs{\phi}_\myw\cup\bs{\phi}_\myz$):
\begin{equation}
    q_{\bs{\phi}}(\myz_{1:T}, \myw | \myx_{1:T}) = q_{\bs{\phi}_{\myw}}(\myw | \myx_{1:T}) \prod_{t=1}^T q_{\bs{\phi}_{\myz}}(\myz_t | \myx_{1:T}, \myw).
\label{eq:dvae_inference}
\end{equation}
% \simonrmk{should it be $\bs{\phi}_\myw$ and $\bs{\phi}_\myz$ as for the generative model?}
The training objective is to maximize the evidence lower bound (ELBO):
\begin{equation}
    \mcal{L}(\bs{\theta},\bs{\phi}; \myx_{1:T}) = \mbb{E}_{q_{\bs{\phi}}(\myz_{1:T},\myw | \myx_{1:T})} \left[ \ln p_{\bs{\theta}}(\myx_{1:T},\myz_{1:T},\myw ) - \ln q_{\bs{\phi}}(\myz_{1:T},\myw | \myx_{1:T}) \right].
\label{eq:dvae_elbo}
\end{equation}
%For the sake of brevity, we omit the dependencies in the following equations.

\begin{figure}[t!]
    \centering
    \includegraphics[width=0.99\linewidth]{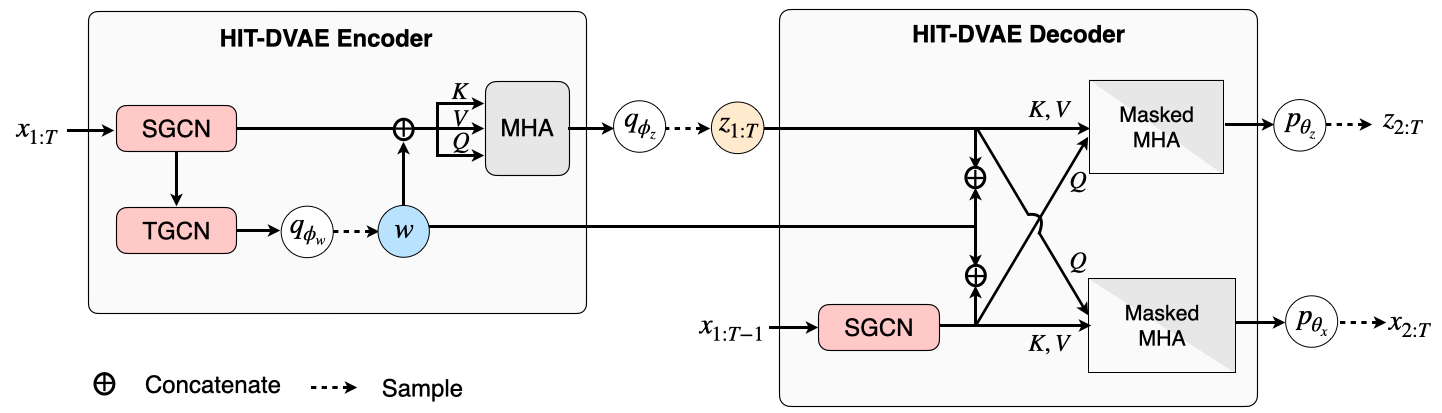} %{figures/pipeline_v2.png}
    \caption{Overview of \method{}. The Encoder (left) inputs the observed sequence $\myx_{1:T}$ to estimate the posterior distribution of the time-dependent $\myz_{1:T}$ and time-independent $\myw$ latent variables. Then the Decoder (right) reconstructs the data and the prior of $\myz$.}% To avoid learning a prior model on $\myx_0$ and $\myz_0$, we start generation at frame 2, this will facilitate the training and do not affect the latter motion generation.
    \label{fig:HIT-DVAE}
\end{figure}

% As for any VAE-based method, we first define the generative model~(\ref{eq:dvae_generation}), also called decoder, and then the inference model~(\ref{eq:dvae_inference}), also called encoder. 
Although the above equations define the probabilistic dependencies between the different random variables, there are plenty of ways of implementing these dependencies. In this paper, we propose to use a hierarchical transformer-based architecture. In our ablation study, we discuss other --perhaps more conventional-- ways of implementing such dependencies, that exhibit lower performance and demonstrate the interest of having both attention and a hierarchical structure, and thus justify the proposed \method{}. Both the encoder and the decoder of the proposed method exploit a spatial graph convolutional network (SGCN) to extract pose features from the raw poses $\myx_t$. We will denote this pose feature extraction operation as $f$, and we will let the encoder and decoder fine-tune their pose extractor leading to $f_\textsc{E}$ and $f_\textsc{d}$, see below for more details. Figure~\ref{fig:HIT-DVAE} shows an overview of our proposed model. Specifically, we employ the transformer architecture~\cite{vaswani2017attention} jointly with GCN-based feature extractors to formulate the inference and generation on the sequential human motion data.

% \noindent \textbf{Generative Model.} 
\subsubsection{Generative Model (\method{} Decoder)} The generation of both $\myx_{1:T}$ and $\myz_{1:T}$ is performed via the attention mechanisms proposed in the original transformer architecture~\cite{vaswani2017attention}. Specifically, the generative model will use multi-head cross-attention, after the GCN-based feature extractor $f_\textsc{d}$. The generative processes of $\myx$ and $\myz$ differ on what variables are used as queries, keys and values in the attention mechanism. The output of the two decoders will be the parameters of the respective probability distributions, defined in~(\ref{eq:dvae_generation}). Both distributions are considered to be Gaussian. While we learn both the mean and covariance matrix of the generative distribution of the latent variable $\myz_t$, the covariance matrix of the observations $\myx_t$ is considered to be the identity as in previous works~\cite{aliakbarian2021contextually,rempe2021humor,guo2020action2motion,petrovich2021action,yuan2020dlow}. In particular we have $p_{\bs{\theta}_{\myx}}(\myx_{t} | \myx_{1:t-1},\myz_t, \myw) = \mathcal{N} (\myx_t;\; \bs{\mu}_{\bs{\mytheta}_{\myx}, t}, \mathbf{I})$ and $p_{\bs{\theta}_{\myz}}(\myz_{t} | \myx_{t-1},\myz_{1:t-1}, \myw) = \mathcal{N} (\myz_t;\; \bs{\mu}_{\bs{\mytheta}_{\myz}, t}, \bs{\Sigma}_{\mytheta_{\myz}, t})$ with 
\begin{align}
\bs{\mu}_{\bs{\mytheta}_{\myx}, t} &= \textrm{MaskedMultiHead} \left(Q_{\bs{\mytheta_{\myx}},t}, K_{\bs{\mytheta_{\myx}}}, V_{\bs{\mytheta_{\myx}}}\right), \\
Q_{\bs{\mytheta_{\myx}},t} &= \left[ \begin{tabular}{c} $\myz_t$ \\ $\myw$ \end{tabular} \right], K_{\bs{\mytheta_{\myx}}} = V_{\bs{\mytheta_{\myx}}} = [f_\textsc{d}(\myx_1), \ldots, f_\textsc{d}(\myx_{T})],\\
\left[ \begin{tabular}{c} 
$\bs{\mu}_{\mytheta_{\myz}, t}$ \\
$\bs{\Sigma}_{\mytheta_{\myz}, t}$ \end{tabular} \right]  &= \textrm{MaskedMultiHead} \left(Q_{\mytheta_{\myz}, t}, K_{\bs{\mytheta_{\myz}}}, V_{\mytheta_{\myz}}\right), \\
Q_{\mytheta_{\myz}, t} &= \left[ \begin{tabular}{c} $f_\textsc{d}(\myx_{t-1})$ \\ $\myw$ \end{tabular} \right], K_{\bs{\mytheta_{\myz}}} = V_{\mytheta_{\myz}} = [\myz_1, \ldots, \myz_{T}],
\end{align}
where a mask is used to prevent $\myz_{t}$ and $\myx_{t}$ from being generated from future latent and observed variables \cite{vaswani2017attention}. Finally, we have $p_{\bs{\theta}_{\myw}}(\myw) = \mathcal{N}(\myw;\mathbf{0},\mathbf{I})$, where $\mathbf{0}$ and $\mathbf{I}$ are the zero vector and the identity matrix of appropriate dimensions. 

% We use two Transformer decoders to generate $\myx_{1:T}$ and $\myz_{1:T}$ respectively. Following the decomposition in Eq. (\ref{eq:dvae_generation}), when generating $\myx_t$, we concatenate $\myz_t$ with $\myw$ as a query and use the previous observations $\myx_{1:T-1}$ as keys and values. Then we concatenate $\myx_{t-1}$ with $\myw$ as a query and $\myz_{1:t-1}$ as keys and values to generate $\myz_t$. Same to the previous works on CVAE based motion generation~\cite{aliakbarian2021contextually,rempe2021humor,guo2020action2motion,petrovich2021action,yuan2020dlow}, we fix the variance of the generated $x_{1:T}$. The generative functions at time $t$ can be wrote as:

% \begin{gather}
%     \bs{\mu}_{\bs{\mytheta}_{\myx}, t} = \text{TransformerDecoder}_{\bs{\mytheta_{\myx}}} (\left[ \begin{tabular}{c} $\myz_t$ \\ $\myw$ \end{tabular} \right] , [\myx_1, ..., \myx_{t-1}]) \\
%     \left[ \begin{tabular}{c} 
%     $\bs{\mu}_{\mytheta_{\myz}, t}$ \\
%     $\bs{\Sigma}_{\mytheta_{\myz}, t}$ \end{tabular} \right]  = \text{TransformerDecoder}_{\bs{\mytheta_{\myz}}} (\left[ \begin{tabular}{c} $\myx_{t-1}$ \\ $\myw$ \end{tabular} \right] , [\myz_1, ..., \myz_{t-1}]) \\
%     \myx_t \sim \mathcal{N} (\myx_t;\; \bs{\mu}_{\bs{\mytheta}_{\myx}, t}, \mathcal{I}) \\
%     \myz_t \sim \mathcal{N} (\myz_t;\; \bs{\mu}_{\bs{\mytheta}_{\myz}, t}, \bs{\Sigma}_{\mytheta_{\myz}, t})
% \end{gather}

\subsubsection{Inference Model (\method{} Encoder)} The inference of the latent variables $\myw$ and $\myz_{1:T}$ from $\myx_{1:T}$ is performed via a temporal GCN and the multi-head self-attention mechanism of the transformer encoder (TE). The extracted pose features are fed into the temporal GCN with $T$ nodes, where each node indicates a time frame, and then into a fully connected (FC) layer to output the mean and variance of $\myw$, namely $\bs{\mu}_{\bs{\myphi_{\myw}}}$ and $\bs{\Sigma}_{\bs{\myphi_{\myw}}}$. Samples are drawn from the corresponding posterior $q_{\bs{\phi}_{\myw}}(\myw | \myx_{1:T}) = \mathcal{N}(\myw;\; \bs{\mu}_{\bs{\myphi_{\myw}}}, \bs{\Sigma}_{\bs{\myphi_{\myw}}})$, concatenated to all pose features and then fed into the transformer encoder such that $q_{\bs{\phi}_\myz}(\myz_t | \myx_{1:T}, \myw) = \mathcal{N} (\myz_t;\; \bs{\mu}_{\myphi_{\myz}, t}, \bs{\Sigma}_{\myphi_{\myz}, t})$ with
\begin{align}
\left[ \begin{tabular}{c} 
$\bs{\mu}_{\myphi_{\myz}, t}$ \\
$\bs{\Sigma}_{\myphi_{\myz}, t}$ \end{tabular} \right]  &= \textrm{MultiHead}\left(Q_{\myphi_{\myz}, t}, K_{\bs{\myphi_{\myz}}}, V_{\myphi_{\myz}}\right), \\
Q_{\myphi_{\myz}, t} &= \left[ \begin{tabular}{ccc} 
    $f_\textsc{E}(\myx_t)$ \\
    $\myw,$ \end{tabular} \right] , K_{\bs{\myphi_{\myz}}} = V_{\myphi_{\myz}} = \left[ \begin{tabular}{ccc} 
    $f_\textsc{E}(\myx_1),$ & $\ldots,$ & $f_\textsc{E}(\myx_T)$  \\
    $\myw$ & $\ldots,$ & $\myw$ \end{tabular} \right].
\end{align}
% where each of the output at index $t$ indicates the distribution parameters of the latent variable $\myz_t$.

% Unlike the previous work on DVAEs that use the recurrent neural networks (e.g. LSTM, GRU) to model the time dependencies, we adopt Transformer~\cite{vaswani2017attention}, a powerful architecture shown impressive performance both on natural language processing~\cite{wolf2020transformers} and computer vision~\cite{han2020survey}. 

% \subsubsection{Probabilistic dependencies via masked attention} In \method{}, the temporal dependencies are implemented via the mask of the attention modules of the transformer decoder and encoder. 

% The attention matrix in a Transformer layer is computed as follows:
% \begin{equation}
%     Att(\bs{Q}, \bs{K}, \bs{V}) = Softmax\left( \bs{\mathcal{M}} \circ \frac{\bs{Q}\bs{K}^T}{\sqrt{d_k}} \right) \bs{V},
% \end{equation}
% where $\bs{Q}, \bs{K}, \bs{V}$ represent the query, key and value, and $d_k$ represents the input feature dimension of query and key. $\bs{\mathcal{M}}$ is the attention mask and $\circ$ denotes the element-wise multiplication. Obviously, an upper triangular mask without diagonal can prevent the model to see the future input. In this case, we can generate the entire sequence of $x_{1:T}$ or $z_{1:T}$ simultaneously. In practice, given an observed sequence with length $T$, we only generate  $x_{2:T}$ and $z_{2:T}$ to bypass the estimation of initial state $x_0$ and $z_0$ as described in~\cite{rempe2021humor}. 

\subsubsection{Feature Extractor on Human Poses} As mentioned above, we extract pose features via a spatial GCN $f$. This strategy was suggested in~\cite{chung2015recurrent}. In \method{}, the spatial GCN is composed of $J$ nodes in the graph, each of them representing a joint in the pose skeleton. While the architecture of the spatial GCN is the same as that of the generative and inference models, these two spatial GCNs are trained separately (i.e., the weights are not shared).

\subsection{Training losses}

\subsubsection{ELBO.} Optimising the evidence lower bound~(\ref{eq:dvae_elbo}) in the case of the proposed \method{}, boils down to (i) minimising the $L_2$ loss on the reconstructed poses while (ii) minimising the KL divergence between the posterior and prior distributions over the latent variables.
Because directly optimising the ELBO would not encourage diversity in our generative model, we inspire from~\cite{mao2021gsps}, and we explicitly generate $K$ motion sequences $\{\hat{\myx}_{1:T}^k\}_{k=1}^K$ and compute the ground-truth reconstruction loss and multi-modal reconstruction loss:
\begin{equation}
    \mathcal{L}_{\textsc{r}} = \min_{k} || \hat{\myx}_{1:T}^k - \myx_{1:T}||^2, \qquad \mathcal{L}_{\textsc{mm}} = \frac{1}{M}\sum_{m=1}^M \min_{k} || \hat{\myx}_{1:T}^k - \myx_{1:T}^m||^2,
    \label{eq:recon}
\end{equation}
where $\myx_{1:T}$ is the ground-truth, and $\myx_{1:T}^m$ are the pseudo-ground truth sequences, which are selected from the training set following the same protocol as in~\cite{mao2021gsps}. In addition to the reconstruction losses, we need to minimise the KL divergence:
\begin{align}
    \mathcal{L}_{\textsc{kl-z}} &= \frac{1}{T}\sum_{t=1}^T D_{KL} ( q_{\bs{\phi}_{\myz}}(\myz_t | \myx_{1:T}, \myw) || p_{\bs{\theta}_{\myz}}(\myz_{t} | \myx_{t-1}, \myz_{1:t-1}, \myw))\\
    \mathcal{L}_{\textsc{kl-w}} &= D_{KL} ( q_{\bs{\phi}_{\myw}}(\myw | \myx_{1:T}) || p_{\bs{\theta}_{\myw}}(\myw)).
\label{eq:loss_kl}
\end{align}
The final evidence lower bound (ELBO) writes:
\begin{equation}
    \mathcal{L}_{\textsc{elbo}} = \lambda_{\textsc{r}}\mathcal{L}_{\textsc{r}} + \lambda_{\textsc{mm}}\mathcal{L}_{\textsc{mm}} + \lambda_{\textsc{kl-z}}\mathcal{L}_{\textsc{kl-z}} + \lambda_{\textsc{kl-w}}\mathcal{L}_{\textsc{kl-w}}.
\end{equation}

% \noindent \textbf{Reconstruction Loss.} We apply an $L_2$ loss on reconstructed poses. Follow~\cite{mao2021gsps}, to encourage diverse generation, we explicitly generate $K$ motion sequences $\{\hat{X_j}\}_{j=1}^K$ and compute ground-truth reconstruction loss and multi-modal reconstruction loss:
% \begin{equation}
%     \mathcal{L}_{r} = \min_{j} || \hat{X}_j - X||^2, \; \mathcal{L}_{mm} = \frac{1}{M}\sum_{m=1}^M \min_{j} || \hat{X}_j - X_m||^2,
% \end{equation}
% where $X$ is the ground truth, and $\{X_m\}_{m=1}^M$ indicate the pseudo ground truths, which are chosen from the training dataset under certain threshold.

% \noindent \textbf{KL Divergence.} Same as other VAE and DVAE models, we have a KL term on latent variables $\myz_{1:T}$ and $\myw$: 
% \begin{align}
%     \mathcal{L}_{KLz} &= \frac{1}{T}\sum_{t=1}^T D_{KL} (q_{\bs{\phi}_{\myw}}(\myw | \myx_{1:T}) || p_{\bs{\theta}_{\myx}}(\myx_{t} | \myx_{1:t-1},\myz_t, \myw))\\
%     \mathcal{L}_{KLw} &= D_{KL} (q_{\bs{\phi}_{\myz}}(\myz_t | \myx_{1:T}, \myw) || p_{\bs{\theta}_{\myw}}(\myw)).
% \label{eq:loss_kl}
% \end{align}
% \simonrmk{There is something wrong in the above equations.}

\subsubsection{Diversity loss.} As suggested by~\cite{yuan2020dlow,mao2021gsps}, we add diversity promoting losses on upper body and lower body:
\begin{equation}
    \mathcal{L}_{\textsc{div}} = \sum_{p\in\{l,u\}}\lambda_{\textsc{div-}p}\frac{2}{K(K-1)} \sum_{k=1}^K \sum_{k'=k+1}^K \exp\left(-\frac{|| \hat{\myx}_{1:T}^{k,p} - \hat{\myx}_{1:T}^{k',p} ||_1}{\alpha_p}\right),
\label{eq:loss_div}
\end{equation}
where $l$ ($u$) indicates the lower (upper) body part and $\alpha_p$ is a normalizing factor.

\subsubsection{Realistic pose loss.} Follow~\cite{mao2021gsps}, we employ three extra losses to penalize for unrealistic poses, $\mathcal{L}_{\textsc{l}}$ for shifting limb length, $\mathcal{L}_{\textsc{a}}$ for aberrant joint angles and $\mathcal{L}_{\textsc{nf}}$ for negative prior pose probability from a pre-trained pose prior model based on normalizing flow~\cite{rezende2015variational,dinh2016density}. 
Altogether, our final training loss writes:
% \begin{small}
\begin{equation}
    \mathcal{L} = \mathcal{L}_{\textsc{elbo}} + \mathcal{L}_{\textsc{div}} + \lambda_{\textsc{l}}\mathcal{L}_{\textsc{l}} + \lambda_{\textsc{a}}\mathcal{L}_{\textsc{a}} + \lambda_{\textsc{nf}}\mathcal{L}_{\textsc{nf}}.
\label{eq:loss_tot}
\end{equation}
% \end{small}

\subsection{\method{} for diverse human motion generation}
The losses above allow to train the proposed \method{} to reconstruct full sequences. In practice, as stated in the problem formulation, \method{} must input $O$ observed frames $\myx_{1:O}$ and generate the following $G$ frames $\myx_{O+1:O+G}$. However, if we use the model trained with ground-truth input over the entire sequence $\myx_{1:T}$ ($T=O+G$) we encounter severe difficulties since after $O$ frames the input distribution changes from the ground-truth to the generated one, and the generation fails. One alternative could be training with ground-truth input until $O$ $\myx_{1:O}$ and then complete with generated data $\hat{\myx}_{O+1:O+G}$. Unfortunately, at the beginning of the training the generated data is pattern-less, and the training diverges. In order to overcome this issue, we use scheduled sampling~\cite{bengio2015scheduled}: we start training only with ground-truth data, and we progressively add more generated frames (randomly) with a proportion starting at $0\%$ and up to $100\%$.

Once our model is trained, we can use it to generate various future motion sequences in arbitrary length. Given $O$ observations in our setting, we can get the posterior of $\myz_{1:O}$ and $\myw$ from the inference model. Then, we can generate the next $G$ frames $\hat{\myx}_{O+1:O+G}$ by recursively applying the generative function on $p_{\bs{\theta}_{\myx}}(\myx_{t} | \myx_{1:t-1},\myz_t, \myw)$ and $ p_{\bs{\theta}_{\myz}}(\myz_{t} | \myx_{t-1},\myz_{1:t-1}, \myw)$. The diversity comes simply from the different samples of $\myz_{O+1:O+G}$ and $\myw$.

% \subsubsection{Schedule Sampling} To be noted that, during training, we have the ground truth $\myx_{1:t-1}$ to generate $\hat{\myx}_t$, whereas we could only have access to $\hat{\myx}_{1:t-1}$ when generating future motion sequences. Follow~\cite{girin2020dynamical}, we use schedule sampling~\cite{bengio2015scheduled}, i.e. alongside with training with ground truth observation, we further train our \method{} decoder by progressively replace $\myx_{1:T-1}$ with $\tilde{{\myx}}_{1:T-1}$ randomly with a proportion going from $0\%$ to $\%100$.

%%%%%%%%%%%%%%%%%%%%%%%%%%%%%%%%%%%%%%%% samples for figs and tabs
% \begin{figure}
% \centering
% \includegraphics[height=6.5cm]{eijkel2}
% \caption{One kernel at $x_s$ ({\it dotted kernel}) or two kernels at
% $x_i$ and $x_j$ ({\it left and right}) lead to the same summed estimate
% at $x_s$. This shows a figure consisting of different types of
% lines. Elements of the figure described in the caption should be set in
% italics,
% in parentheses, as shown in this sample caption. The last
% sentence of a figure caption should generally end without a full stop}
% \label{fig:example}
% \end{figure}

% \begin{align}
%   \psi (u) & = \int_{0}^{T} \left[\frac{1}{2}
%   \left(\Lambda_{0}^{-1} u,u\right) + N^{\ast} (-u)\right] dt \; \\
% & = 0 ?
% \end{align}

%% file: 4-Experiments.tex
\section{Experiments}

%%%%%%%%%%%%%%%%%%%%%%%%%%%%%%%%%%%%%%%%
\subsection{Evaluation Protocols} %dlow
\label{sec:eval_metric}
\input{fig_vis}
\subsubsection{Datasets}
Following~\cite{mao2021gsps,yuan2020dlow}, we train and evaluate our methods on Human3.6M~\cite{ionescu2013human3} and HumanEva-I~\cite{sigal2006humaneva} dataset: \textbf{Human3.6M} is the most commonly used dataset for motion tasks. It contains 7 actors (S1,5,6,7,8,9,11) performing 15 annotated actions recorded at 50 Hz. Human pose is represented in 32 skeletons, while we follow~\cite{mao2021gsps} to use 17 of the skeletons in our training and all testing implementations, and we use S1,5,6,7,8 as training set and the other two subjects as test set.
\textbf{HumanEva} contains 5 actions (Box, Gesture, Jog, ThrowCatch, Walking) performed by 3 actors, recorded at 60 Hz. Each pose is represented by 15 skeletons. 
For both datasets, we remove the global translation and set the root joint as zero.

\input{tab_gen_eva}

\subsubsection{Explicit evaluation metrics}
Following \cite{mao2021gsps,yuan2020dlow}, we evaluate error and diversity of our results with the following metrics, calculating directly on the joint locations of poses:
i) Average Pairwise Distance (APD): average $L2$ distance of all pairs among the generated sequences: $\frac{1}{K(K-1)}\sum_{i=1}^K \sum_{j\neq i}^K \|\hat{\mathbf{x}}^i_{O+1:O+G} - \hat{\mathbf{x}}^j_{O+1:O+G}\|_2$. APD measures the capacity of the model to generate diverse samples without considering their quality.
ii) Average Displacement Error (ADE): $L2$ distance between the ground truth and the 'best' generated sample among all, taking the average over all frames of the sequence:$\frac{1}{G}\min_k \|\hat{\mathbf{x}}^k_{O+1:O+G} - \mathbf{x}_{O+1:O+G}\|$. Here 'best' means the one which is closest to the ground truth. ADE evaluates the upper bound of the generation quality of the model among all the generation results.
iii) Final Displacement Error (FDE): Similar to ADE, FDE evaluates the distance between the ground truth and the best sample, but just on the final frame instead of the whole sequence: $\min_k \|\hat{\mathbf{x}}^k_{O+G} - \mathbf{x}_{O+G}\|$. 
iv) Multi-Modal ADE (MMADE) and Multi-Modal FDE (MMFDE): multi-modal version of ADE and FDE. 

The MPJPE based metrics are widely used for evaluating the quality of generated motion, considering the diversity of the generated data and accurancy of the best generated one. While the shortcoming of them is obvious. On one hand, when we use the generator to generate actions, we could not always have the groundtruth to judge which generation is the best one; on the other hand, as described in Sec~\ref{sec:realated_work}, generating one best example is what deterministic motion prediction aims at, while stochastic methods should generate various good samples. For example, a batch of generated motions where only one sample is exact while the others are all super crazy will result in a large APD and tiny errors which seems perfect numerically, but this is certainly not a good generation we want. So just considering the above metrics is not comprehensively and proper.

Thus, we take two solutions: 1) instead of just evaluating ii-iv) on the best generated sample, we also evaluate these criteria on the 'medium' example, which holds the medium distance to the groundtruth among all the generated samples. 2) beside of this explicit measurement based on poses, we also consider implicit measurements based on a pretrained action classifier, as described below.
%\wen{TODO: modify the notations in the equations}

\subsubsection{Implicit evaluation metrics} 
Following \cite{petrovich2021action,guo2020action2motion}, we use a GRU-based action classifier pre-trained on real data to evaluate the quality of generated data by: \\
i) calculating Recognition Accuracy (Acc) of the classifier on generated data to evaluate if the generated data could be recognized as the correct action class.\\
ii) extracting features of the generated data and real data respectively by the action classifier, and calculate the Frechet Inception Distance (FID) of these two distributions to evaluate the overall quality of generated data.
For the two datasets, we trained a classifier for each of them on their training splits.

% \noindent\textbf{Downstream task: data augmentation for motion prediction}
% Beside of the above metrics, we also tested the effectiveness of the generated data when serving as augmented data of state of the art single person motion prediction moethods~\cite{mao2019learning,mao2020history}, and compare the performance of different generating methods.

%%%%%%%%%%%%%%%%%%%%%%%%%%%%%%%%%%%%%%%%
\subsection{Implementation details} 
We set the dimension of $\myz_t$ to 16 and $\myw$ to 32, and employ the same GCN architecture described in~\cite{mao2019learning}. We use 1 GCN block with hidden size of 8 for spatial GCN and 4 GCN blocks with hidden size of 64 for temporal GCN. For the Transformer encoder and the decoder for generating $\myz_t$, we set the input feature dimension to 64, with 4 multi-head, followed by a FC layer with dimension of 256, whereas for the Transformer decoder to generate $\myx_t$, we set those parameters to 256, 4, 1024 respectively. 

We generate $K=50$ samples for each observation. We train the model for 500 epochs with 1000 training samples per epoch, using Adam optimizer, and set learning rate to 0.001, batch size to 64 for HumanEva and 32 for H3.6M. We applied a linear KL annealing~\cite{sonderby2016ladder} for the first 20 epochs to warm-up the latent space, then we take 80 epochs to increase the probability of schedule sampling from 0 to 1. For HumanEva, we train with a sequence length of 75, where the inference of $\myw$ only takes 15 frames with a random start point. The weights of different loss terms $(\lambda_r, \lambda_{mm}, \lambda_{d, l}, \lambda_{d, u}, \lambda_{l}, \lambda_{a}, \lambda_{nf}, \lambda_z , \lambda_w)$ and the normalizing factors $(\alpha_l, \alpha_u)$ are set to (10, 5, 0.1, 0.2, 100, 1, 0.001, 0.5, 0.1) and (15, 50). For H3.6M,  we train with a sequence length of 125, where $\myw$ is inferred from 25 frames. The weights of different loss terms and the normalizing factors are set to (20, 10, 0.1, 0.2, 100, 1, 0.01, 0.5, 0.1) and (100, 300) respectively.

%%%%%%%%%%%%%%%%%%%%%%%%%%%%%%%%%%%%%%%%

\input{tab_gen_h36m}
\subsection{Quantitative results}
We evaluate the generated motions on HumanEva-I and Human3.6m dataset by the explicit and implicit metrics described in Sec~\ref{sec:eval_metric}, and found that our methods outperforms the state of the art methods on most of the evaluation metrics.

\subsubsection{HumanEva-I}
As shown in Tab~\ref{table:tab_res_eva}, our method achieves comparable results with state of the art on explicit evaluation of diversity (APD) and errors of 'best' sample (ADE$_b$, FDE$_b$, MMADE$_b$, MMFDE$_b$). As discussed in Sec~\ref{sec:eval_metric}, we know that just considering these errors and diversity is not reliable, because these erros just evaluate the best sample among all generations and APD just evaluates diversity without considering the generation quality. And we should note that APD is not always better for larger values, because although we want the generated data to have diversity, crazy large diversity represents that some of the generated samples might totally fail and the quality of generation is not guarantied.
So in order to comprehensively measure the performance of generated data, we test the error of 'medium' samples, action-recognition based accuracy ACC and feature-based FID scores on our data, and also on other code-released state of the art methods~\cite{mao2021gsps,yuan2020dlow}.

We find that our method achieves significantly better results than other state of the art methods on the implicit metrics ACC and FID, which means that our generated data is better in feature distribution, and could generate more reasonable results that could be recognized as the right action. And our method is also clearly better on errors of medium sample (ADE$_m$, FDE$_m$, MMADE$_m$, MMFDE$_m$), which indicates higher stability of our overall generation quality.\\

\subsubsection{Human3.6M}
Similar conclusion could be drawn for Human3.6M dataset, as shown in Tab~\ref{table:tab_res_h36m}. 
When training the action classifier for Human3.6M dataset, 
instead of training on all the 15 action labels, we group the 15 actions into 5 groups. This is because that some actions in Human3.6M dataset are not with much difference, so it is is not suitable for training the action classier. For example, we could not see the difference between 'eating' and 'smoking' with just the skeletons of the person. With this grouping, the average classification accuracy on real data of our classifier increases from 48.1\% to 85.5\%. Note that even on the 15-action classifier with low real-data-accuracy, our method still has higher Acc and lower FID comparing with other state of the art methods. While we report the 5-group classifier here because we believe that a better classifier is more reliable for calculating accuracy and extracting features for FID. More details about the action classifier could be found in supplementary material.%\wen{In supp: explain the details of the action classifier models, like model structure, parameters, action splitting, results of ours vs gsps vs delow on 15-splits.}\\

% %%%%%%%%%%%%%%%%%%%%%%%%%%%%%%%%%%%%%%%%

\input{tab_abla}
\subsection{Ablation study}
Tab~\ref{table:tab_abla} shows ablation studies on our method with different architecture design. We bold the best results and underlined the second best ones. We could find that, without schedule sampling, our method tends to generate more diverse results, but with worse quality either on explicit metrics and implicit metrics. 
%The benefit of introducing attention is not evident on the HumanEva-I, but when we l
Looking in the results of Human3.6M, the use of attention mechanism brings higher generation quality on explicit measurement. The global time-independent variable $\myw$ bring more diversity on both two datasets (see the results w/o $\myw$). When we consider a vanilla DVAE model (w/o Att. \& $\myw$), without the hierarchical transformer (HIT) architecture, it is very likely to collapse to a static state on the sequential latent space, which leads to moderate generation quality, and much worse diversity. The final setting of \method{} we used performs good on almost all the metrics and is a balance of different evaluations.

% %%%%%%%%%%%%%%%%%%%%%%%%%%%%%%%%%%%%%%%%
% \subsection{Data augmentation for motion prediction}
% We evaluate the generated motions by taking the generated motion as augmented data of the original dataset, and use them to train the motion prediction tasks.\\
% note: For H36m dataset, the original motion prediction models are trained with 22 joints, while to maintain the consistency with the generation models, we retrain all these prediction models on 17 joints.\\
% Tab~\ref{table:tab_aug}
% \input{tables/tab_data_aug}

%%%%%%%%%%%%%%%%%%%%%%%%%%%%%%%%%%%%%%%%
\subsection{Qualitative results}
To qualitatively evaluate our generated results, we visualize various generating samples of our methods in Fig~\ref{fig:vis} comparing with other state of the art methods. we could see that other methods either generates very similar samples for all the generations, either result in some weird actions, while our method performs well on all the generations, with diverse but always reasonable results. More visualisations could be found in supplementary material.

%% file: fig_vis.tex
\begin{figure}[th]
    \centering
    \includegraphics[width=0.99\linewidth]{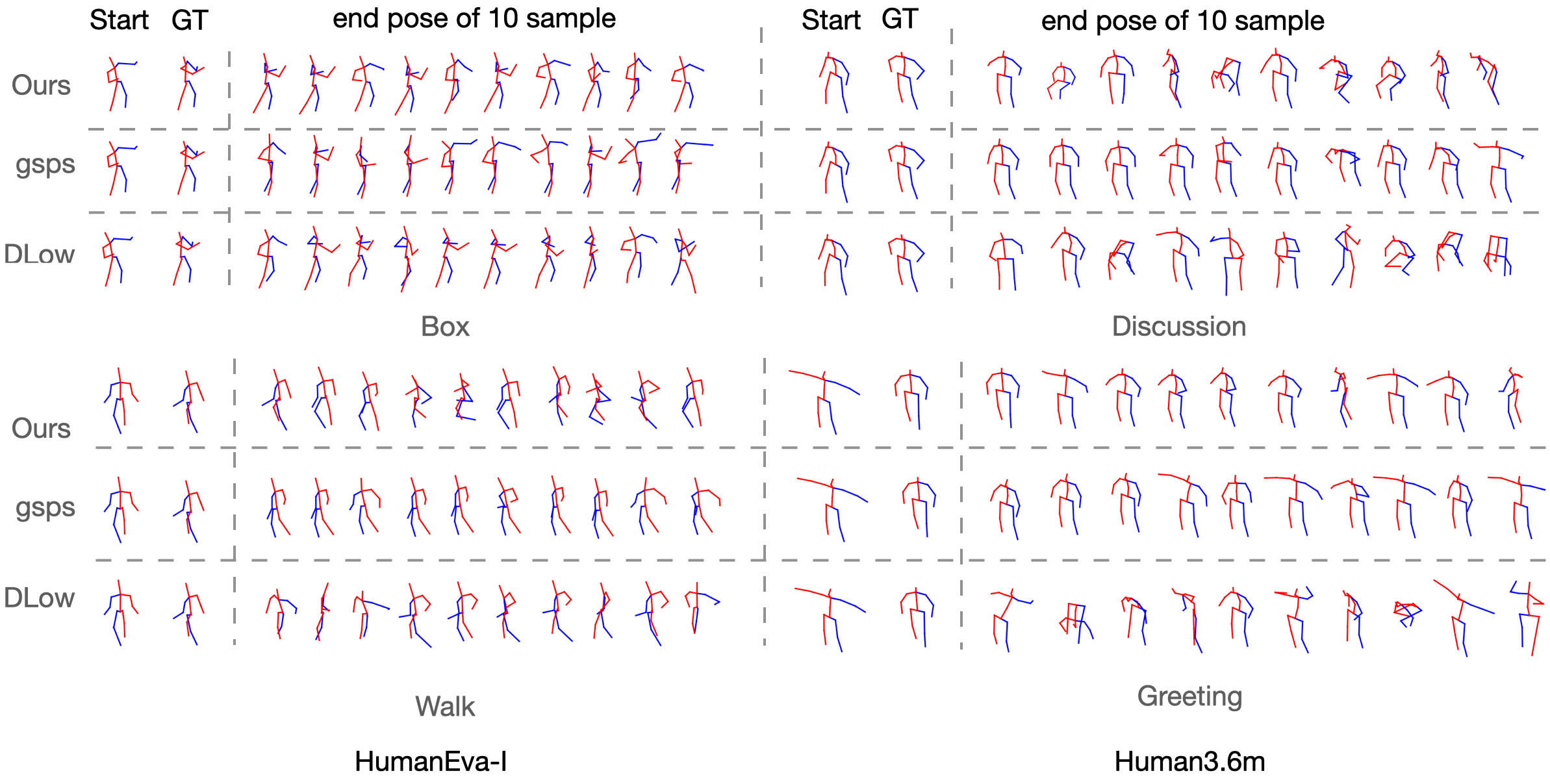}
    \vspace{-3mm}
    \caption{{\bf Qualitative visualization} on four different actions of the HumanEva and Human3.6M datasets. For each case we show 'Start', which is the last observed frame and the subsequent frames correspond the the last  frames of 10 different generated sequences. GT is the ground truth last frame. Note that gsps and DLow highly diverge from that last GT frame, while our approach, while generating different alternatives, they all keep the essence of the particular action.}
    \label{fig:vis}
\end{figure}

%% file: tab_gen_eva.tex
%%%%%%%%%%%%%%%%%%%%%%%%%%%%%%%%%%%%%%%%%%%%%%%%%%%%%%%%%%%%%%%%%%%%%%%%%%%%%%%%%%%%%%%%%%%%%%%%%%% eva
\setlength{\tabcolsep}{4pt}
\begin{table}[t]
\begin{center}
\caption{\textbf{Results on HumanEva-I.} ``Real data'' means real motion in testing set, shown the theoretical upper bounds of accuracy on generation methods. The suffix 'b' or 'm' represents the best/medium metrics. $\uparrow$ ($\downarrow$) means higher (lower) is better. $\dag$ indicates results taken from DLow, $\star$ indicates results obtained by using the official code repository.}
\label{table:tab_res_eva}
\resizebox{0.9\textwidth}{!}{\setlength{\tabcolsep}{0.8mm}{
\begin{tabular}{l|cc|c|cccc|cccc}
\toprule%\noalign{\smallskip}
& \makecell[c]{Acc\\(\%) $\uparrow$} & \makecell[c]{FID\\$\downarrow$} & \makecell[c]{APD\\(m) $\uparrow$} & \makecell[c]{ADEb\\(m) $\downarrow$} & \makecell[c]{FDEb\\(m) $\downarrow$} & \makecell[c]{MMADEb\\(m) $\downarrow$} & \makecell[c]{MMFDEb\\(m) $\downarrow$} & \makecell[c]{ADEm\\(m) $\downarrow$} & \makecell[c]{FDEm\\(m) $\downarrow$} & \makecell[c]{MMADEm\\(m) $\downarrow$} & \makecell[c]{MMFDEm\\(m) $\downarrow$}\\
% \noalign{\smallskip}
\midrule
% \noalign{\smallskip}
Real data & 88.3 & - & - &- &- &- &- &- &- & - & -\\
\midrule
ERD$\dag$~\cite{fragkiadaki2015recurrent} & - &- & 0 & 0.382 & 0.461 & 0.521 & 0.595 &- &- & - & - \\
acLSTM$\dag$~\cite{li2017auto} & - &- &  0 & 0.429 & 0.541 & 0.530 & 0.608 &- &- & - & -\\
Pose-Knows$\dag$~\cite{walker2017pose} & - &- & 2.308 & 0.269 & 0.296 & 0.384 & 0.375 &- &- & - & -\\
MT-VAE$\dag$~\cite{yan2018mt} &-  &- & 0.021 & 0.345 & 0.403 & 0.518 & 0.577 &- &- & - & -\\
HP-GAN$\dag$~\cite{barsoum2018hp}  & - &- & 1.139 & 0.772 & 0.749 & 0.776 & 0.769 &- &- & - & -\\
BoM$\dag$~\cite{bhattacharyya2018accurate} & - &- & 2.846 & 0.271 & 0.279 & 0.373 & 0.351 &- &- & - & -\\
GMVAE$\dag$~\cite{dilokthanakul2016deep}  & - &- & 2.443 & 0.305 & 0.345 & 0.408 & 0.410 &- &- & - & -\\
DeLiGAN$\dag$~\cite{gurumurthy2017deligan} & - &- & 2.177 & 0.306 & 0.322 & 0.385 & 0.371 &- &- & - & -\\
DSF$\dag$~\cite{yuan2019diverse} & - &- & 4.538 & 0.273 & 0.290 & 0.364 & 0.340 &- &- & - & -\\
%cvae & 63.3 & & & & \\
%delow\dag~\cite{yuan2020dlow} & 60.6 & & 4.855 & 0.251 & 0.268 & 0.362 & 0.339 &- &- & - & -\\
%gsps\dag~\cite{mao2021gsps} & 66.7 & & 5.825 & 0.233 & 0.244 & 0.343 & 0.331 &- &- & - & - \\
DLow$\star$~\cite{yuan2020dlow} & 52.7  & 3.472 & 4.853 & 0.248 & 0.262 & 0.361 & 0.337 & 0.577 & 0.717 & 0.646 & 0.753\\
gsps$\star$~\cite{mao2021gsps}  & 51.6  & 1.604 & \textbf{5.825} & \textbf{0.233} & \textbf{0.244} & 0.343 & 0.311 & 0.686 & 0.794 & 0.735 & 0.825\\
\midrule
%ours div0.2 0.4& 70.6 & 69.9 & 33.398 & 0.128 & 4.730 & 0.271 & 0.249 & \textbf{0.330} & \textbf{0.290} & \textbf{0.572} & \textbf{0.658} & \textbf{0.605} & \textbf{0.673} \\
\method{}  & \textbf{72.6} & \textbf{0.089} & 4.721 & 0.282 & 0.261 & \textbf{0.335} & \textbf{0.290} & \textbf{0.579} & \textbf{0.665} & \textbf{0.610} & \textbf{0.683} \\
\bottomrule
\end{tabular}}}
\end{center}
\end{table}
\setlength{\tabcolsep}{1.4pt}

%% file: tab_gen_h36m.tex
%%%%%%%%%%%%%%%%%%%%%%%%%%%%%%%%%%%%%%%%%%%%%%%%%%%%%%%%%%%%%%%%%%%%%%%%%%%%%%%%%%%%%%%%%%%%%%%%%%% h36m

\setlength{\tabcolsep}{4pt}
\begin{table}[t]
\begin{center}
\caption{\textbf{Results on Human3.6M.} ``Real data'' means real motion in testing set, shown the theoretical upper bounds of accuracy on generation methods. The suffix 'b' or 'm' represents the best/medium metrics. $\uparrow$ ($\downarrow$) means higher (lower) is better. $\dag$ indicates results taken from DLow, $\star$ indicates results obtained by using the official code repository.}
\label{table:tab_res_h36m}
\resizebox{0.9\textwidth}{!}{\setlength{\tabcolsep}{0.8mm}{
\begin{tabular}{l|cc|c|cccc|cccc}
\toprule%\noalign{\smallskip}
% & ACCUr \uparrow &ACCUs \uparrow & APD  \uparrow & ADEb \downarrow & FDEb \downarrow & \makecell[c]{MM\\ADEb}\downarrow & \makecell[c]{MM\\FDEb} \downarrow & ADEm \downarrow & FDEm \downarrow & \makecell[c]{MM\\ADEm}\downarrow & \makecell[c]{MM\\FDEm}\downarrow\\
& \makecell[c]{Acc\\(\%) $\uparrow$} & \makecell[c]{FID\\$\downarrow$} & \makecell[c]{APD\\(m) $\uparrow$} & \makecell[c]{ADEb\\(m) $\downarrow$} & \makecell[c]{FDEb\\(m) $\downarrow$} & \makecell[c]{MMADEb\\(m) $\downarrow$} & \makecell[c]{MMFDEb\\(m) $\downarrow$} & \makecell[c]{ADEm\\(m) $\downarrow$} & \makecell[c]{FDEm\\(m) $\downarrow$} & \makecell[c]{MMADEm\\(m) $\downarrow$} & \makecell[c]{MMFDEm\\(m) $\downarrow$}\\
%\noalign{\smallskip}
\midrule
%\noalign{\smallskip}
Real data & 85.5 & - & - &- &- &- &- &- &- & - & -\\
\midrule
ERD$\dag$~\cite{fragkiadaki2015recurrent} &- &-  & 0  & 0.722 & 0.969 & 0.776 & 0.995 &- &- & - & -\\
acLSTM$\dag$~\cite{li2017auto} &  - &-  & 0  & 0.789 & 1.126 & 0.849 & 1.139 &- &- & - & -\\
Pose-Knows$\dag$~\cite{walker2017pose} & - &-  & 6.723  & 0.461 & 0.560 & 0.522 & 0.569 &- &- & - & -\\
MT-VAE$\dag$~\cite{yan2018mt} &- &-  & 0.403  & 0.457 & 0.595 & 0.716 & 0.883 &- &- & - & -\\
HP-GAN$\dag$~\cite{barsoum2018hp} &- &- & 7.214  & 0.858 & 0.867 & 0.847 & 0.858 &- &- & - & -\\
BoM$\dag$~\cite{bhattacharyya2018accurate} &- &- & 6.265  & 0.448 & 0.533 & 0.514 & 0.544 &- &- & - & -\\
GMVAE$\dag$~\cite{dilokthanakul2016deep} & - &- & 6.769  & 0.461 & 0.555 & 0.524 & 0.566 &- &- & - & -\\
DeLiGAN$\dag$~\cite{gurumurthy2017deligan} & - &-  & 6.509  & 0.483 & 0.534 & 0.520 & 0.545 &- &- & - & -\\
DSF$\dag$~\cite{yuan2019diverse} &- &- & 9.330  & 0.493 & 0.592 & 0.550 & 0.599 &- &- & - & -\\
%cvae & 63.3 & & & & \\
%delow\dag~\cite{yuan2020dlow} &  & & 11.741 & 0.425 & 0.518 & 0.495 & 0.531 &- &- & - & -\\
%gsps\dag~\cite{mao2021gsps} &  & & 14.757 & 0.389 & 0.496 & 0.476  & 0.525 &- &- & - & -\\
DLow$\star$~\cite{yuan2020dlow} & 65.9 & \textbf{1.412} & 11.741 & 0.425 & 0.518 & 0.495 & 0.531 & 0.896 & 1.284 & 0.948 & 1.289\\
gsps$\star$~\cite{mao2021gsps} & 65.0 & 2.030 & \textbf{14.757} &\textbf{ 0.389} & \textbf{0.496} & \textbf{0.476} & 0.525 & 1.013 & 1.372 & 1.065 & 1.381\\
\midrule
ours & \textbf{70.0} & 1.708 & 8.942 & 0.472 & 0.505 & 0.497 & \textbf{0.514} & \textbf{0.804} & \textbf{1.034} & \textbf{0.812} & \textbf{1.028}\\
\bottomrule
\end{tabular}}}
\end{center}
\end{table}
\setlength{\tabcolsep}{1.4pt}

%% file: tab_abla.tex
\setlength{\tabcolsep}{4pt}
\begin{table}
\begin{center}
\caption{\textbf{Ablation Study} on HumanEva-I and Human3.6M. 'w/o SS' means without scheduled sampling, 'w/o Att.' means using an LSTM instead of transformer, 'w/o $\myw$' means without using the time-independent latent variable $\myw$. 'w/o Att. \& $\myw$ ' means without attention and no use of $\myw$.}
\label{table:tab_abla}
%\rowcolors{3}{gray!15}{white}
\resizebox{0.9\textwidth}{!}{\setlength{\tabcolsep}{0.8mm}{
\begin{tabular}{l|cc|c|cccc|cccc}
\toprule%\noalign{\smallskip}
Architecture & \makecell[c]{ACC\\(\%) $\uparrow$} & \makecell[c]{FID\\$\downarrow$} & \makecell[c]{APD\\(m) $\uparrow$} & \makecell[c]{ADEb\\(m) $\downarrow$} & \makecell[c]{FDEb\\(m) $\downarrow$} & \makecell[c]{MMADEb\\(m) $\downarrow$} & \makecell[c]{MMFDEb\\(m) $\downarrow$} & \makecell[c]{ADEm\\(m) $\downarrow$} & \makecell[c]{FDEm\\(m) $\downarrow$} & \makecell[c]{MMADEm\\(m) $\downarrow$} & \makecell[c]{MMFDEm\\(m) $\downarrow$}\\
% \noalign{\smallskip}
\midrule
% \noalign{\smallskip}
\multicolumn{12}{c}{HumanEva-I}\\
\midrule
\method{}  & 72.6 & \textbf{0.089} & \underline{4.721} & \underline{0.282} & \underline{0.261} & \textbf{0.335} & \textbf{0.290} & {0.579} & {0.665} & {0.610} & {0.683} \\
w/o SS & 69.6 & 0.359 & \textbf{4.777} & 0.314 & 0.300 & 0.358 & 0.315 & 0.596 & 0.708 & 0.624 & 0.727\\
w/o Att. & 72.9 & \underline{0.264} & 3.921 & \textbf{0.265} & \textbf{0.243} & \underline{0.348} & \underline{0.295} & \textbf{0.510} & \textbf{0.604} & \textbf{0.569} & \textbf{0.650}\\
w/o $\myw$ & \underline{74.0} & 0.306 & 4.244 & 0.287 & 0.263 & 0.360 & 0.303 & \underline{0.535} & 0.662 & \underline{0.591} & 0.689\\
w/o Att. \& $\myw$ & \textbf{76.5} & 1.262 & 0.023 & 0.538 & 0.621 & 0.594 & 0.660 & 0.538 & \underline{0.622} & 0.595 & \underline{0.662} \\
\midrule
% \noalign{\smallskip}
\multicolumn{12}{c}{Human3.6M}\\
\midrule
\method{} & {70.0} & 1.708 & \underline{8.942} & \textbf{0.472} & \textbf{0.505} & \textbf{0.497} & \textbf{0.514} & \underline{0.804} & \underline{1.034} & \underline{0.812} & \underline{1.028} \\
w/o SS & 65.0 & 1.751 & \textbf{10.339} & \underline{0.477} & \underline{0.513} & \underline{0.501} & \underline{0.519} & 0.867 & 1.108 & 0.877 & 1.101\\
w/o Att. & \underline{70.5} & \underline{1.475} & 7.189 & 0.537 & 0.569 & 0.555 & 0.567 & 0.805 & 1.060 & 0.826 & 1.056\\
w/o $\myw$ & \textbf{71.1} & 1.565 & 6.249 & 0.528 & 0.582 & 0.557 & 0.590 & \textbf{0.735} & \textbf{0.942} & \textbf{0.753} & \textbf{0.942}\\
w/o Att. \& $\myw$ & 64.2 & \textbf{0.659} & 2.688 & 0.752 & 0.892 & 0.787 & 0.896 & 0.897 & 1.184 & 0.934 & 1.186\\
\bottomrule
\end{tabular}}}
\end{center}
\end{table}
\setlength{\tabcolsep}{1.4pt}

%% file: 5-Appendix.tex
%%%%%%%%%%%%%%%%%%%%%%%%%%%%%%%%%%%%%%%%
\appendix

% %%%%%%%%%%%%%%%%%%%%%%%%%%%%%%%%%%%%%%%%
\section{Probabilistic Dependencies via Masked MHA}
The temporal dependencies are implemented via the mask of the attention modules of the transformer decoder and encoder. The attention matrix in a Transformer layer, and is computed as follows:
\begin{equation}
    \textrm{Att}(\bs{Q}, \bs{K}, \bs{V}) = \textrm{Softmax}\left( \bs{\mathcal{M}} \circ \frac{\bs{Q}\bs{K}^T}{\sqrt{d_k}} \right) \bs{V},
\end{equation}
where $\bs{Q}, \bs{K}, \bs{V}$ represent the query, key and value, and $d_k$ represents the input feature dimension of query and key. $\bs{\mathcal{M}}$ is the attention mask and $\circ$ denotes the element-wise multiplication. Obviously, an upper triangular mask without diagonal can prevent the model to see the future input. In this case, we can generate the entire sequence of ${\bf x}_{1:T}$ or ${\bf z}_{1:T}$ simultaneously. In practice, given an observed sequence with length $T$, we only generate  ${\bf x}_{2:T}$ and ${\bf z}_{2:T}$ to bypass the estimation of initial state ${\bf x}_0$ and ${\bf z}_0$. 

\input{fig_mask}

Fig.~\ref{fig:mask} shows three cases of probabilistic dependencies when using different mask in the Transformer layer. Note that Fig.~\ref{fig:mask} (a) is a non-causal situation, thus we can not generate future motion via this dependencies. The mask in Fig.~\ref{fig:mask} (c) will make the attention computed only on one element, thus the attention mask is meaningless in this case. We choose the mask shown in Fig.~\ref{fig:mask} (b) in our proposed HiT-DVAE.

% %%%%%%%%%%%%%%%%%%%%%%%%%%%%%%%%%%%%%%%%
\section{Pseudo-code for HiT-DVAE}
Here, we provide the pseudo-code for HiT-DVAE in training and generation:
\begin{algorithm}[h]
\caption{HiT-DVAE in training}
\begin{algorithmic}
\Inputs{}\vspace{-4mm}
\State{$\triangleright$ Observation on human sequence $\myx_{1:T}$}
\For{epo in epochs}
\State{\textbf{Inference:}}
\State{$\triangleright$ Compute posterior of $\myw$ and sample $ \myw \sim q_{\bs{\phi}_{\myw}}(\myw | \myx_{1:T}) =  \mathcal{N}(\myw;\; \bs{\mu}_{\bs{\myphi_{\myw}}}, \bs{\Sigma}_{\bs{\myphi_{\myw}}})$ }
\State{$\triangleright$ Compute posterior $\myz_{1:T}$ and sample $\myz_t \sim q_{\bs{\phi}_\myz}(\myz_t | \myx_{1:T}, \myw) = \mathcal{N} (\myz_t;\; \bs{\mu}_{\myphi_{\myz}, t}, \bs{\Sigma}_{\myphi_{\myz}, t})$ for $t=1,...,T$ }
\State{\textbf{Generation:}}
\State{$\triangleright$ Compute the distribution of $\myx_{2:T}$ via $p_{\bs{\theta}_{\myx}}(\myx_{t} | \myx_{1:t-1},\myz_t, \myw) = \mathcal{N} (\myx_t;\; \bs{\mu}_{\bs{\mytheta}_{\myx}, t}, \mathbf{I})$ for $t=2, ..., T$}
\State{$\triangleright$ Compute the prior of $\myz_{2:T}$ via  $p_{\bs{\theta}_{\myz}}(\myz_{t} | \myx_{t-1},\myz_{1:t-1}, \myw) = \mathcal{N} (\myz_t;\; \bs{\mu}_{\bs{\mytheta}_{\myz}, t}, \bs{\Sigma}_{\mytheta_{\myz}, t})$ for $t=2, ..., T$}
\State{\textbf{Compute loss and optimize via Adam}}
\EndFor
\end{algorithmic}
\label{algo:vem-enhancement}
\end{algorithm}

\vspace{9mm}

\begin{algorithm}[h]
\caption{HiT-DVAE in generation}
\begin{algorithmic}
\Inputs{}\vspace{-4mm}
\State{$\triangleright$ Observation on human sequence $\myx_{1:O}$}
\Init{}\vspace{-4mm}
\State{$\triangleright$ Compute posterior of $\myw$ and $\myz_{1:O}$}
\For{t in range(O+1, O+G)}
\State{$\triangleright$ Generate $\hat{\myz}_t$ via $\myz_t \sim p_{\bs{\theta}_{\myz}}(\myz_{t} | \hat{\myx}_{t-1},\myz_{1:t-1}, \myw)$}
\State{$\triangleright$ Generate $\hat{\myx}_t$ via $\myx_t \sim p_{\bs{\theta}_{\myx}}(\myx_{t} | \myx_{1:O}, \hat{\myx}_{O+1:t-1},\hat{\myz}_t, \myw) $}
\EndFor
\Outputs{}\vspace{-4mm}
\State{$\triangleright$ Generated human motion sequence $\hat{\myx}_{O+1:O+G}$}
\end{algorithmic}
\label{algo:vem-enhancement}
\end{algorithm}
\vspace{-9mm}

% %%%%%%%%%%%%%%%%%%%%%%%%%%%%%%%%%%%%%%%%
% \vspace{-5mm}
\section{Action-classifier}
As explained in Sec.~4.1 and Sec.~4.3 of the main paper, we trained a RNN-based classifier to calculate the implicit evaluation metrics ACC and FID following~\cite{guo2020action2motion,petrovich2021action}.
The classifier we used is build upon 2 simple GRU layers with hidden size of 128. 
When training on Human3.6M dataset, we found that some action classes do not differ much from each other, which makes it difficult to train a good classifier. As our goal is to have a classifier which offers good feature, we believe the classifier with low accuracy on real data is not reliable enough, so we group the 5 similar actions, 
%due to visualisations and testing results of the classifier on real data, 
and trained the classifier on these 5 groups instead of the 15 original classes. The groups of actions are detailed in Tab~\ref{table:group_15_class}.\\
Note that even on the classifier trained on the 15 original classes, our method still performs better than others, as shown in Tab.~\ref{table:res_15_class}.

\input{tab_supp_act_class_h36m}

% % %%%%%%%%%%%%%%%%%%%%%%%%%%%%%%%%%%%%%%%%
% \section{Data augmentation}
% Beside of the above metrics, we also tested the effectiveness of the generated data when serving as augmented data of state of the art single person motion prediction moethods~\cite{mao2019learning,mao2020history}, and compare the performance of different generating methods.
% We evaluate the generated motions by taking the generated motion as augmented data of the original dataset, and use them to train the motion prediction tasks. 
% Note that for H36m dataset, the original motion prediction models are trained with 22 joints, while to maintain the consistency with the generation models, we retrain all these prediction models on 17 joints.

% \biermk{TODO: generate data before the weekend!}
% \input{tables/tab_data_aug}

% %%%%%%%%%%%%%%%%%%%%%%%%%%%%%%%%%%%%%%%%
% \section{More Visualisation Results}
% As shown in the main paper, we qualitatively compare our methods with the state-of-the-art methods, gsps~\cite{mao2021gsps} and DLow~\cite{yuan2020dlow}, by showing the end poses of 10 generated motion sequences. In the attached video, we further compare our results with them by visualizing 10 generation samples for each method. For better visualisation, we highlight the 'unrealistic' generated motions which are evidently not corresponded to the action label and not similar with the real motion, based on human study.
% We can see that our method generates more realistic motion sequences and keep the diversity, while others are less stable.

%% file: fig_mask.tex
\begin{figure*}[htbp]
    \centering
    \begin{tabular}{ccc}
    \includegraphics[width=0.25\linewidth]{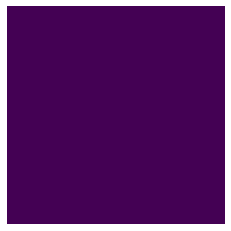} &
    \includegraphics[width=0.25\linewidth]{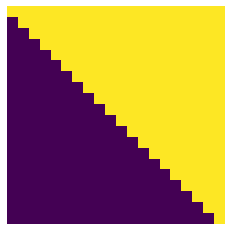} &
    \includegraphics[width=0.25\linewidth]{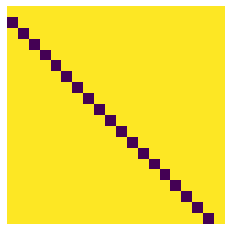} \\
    {(a) $p_{\bs{\theta}_{\myx}}(\myx_{t} | \myx_{1:T},\myz_t, \myw)$ } & 
    {(b) $p_{\bs{\theta}_{\myx}}(\myx_{t} | \myx_{1:t-1},\myz_t, \myw)$ } & 
    {(c) $p_{\bs{\theta}_{\myx}}(\myx_{t} | \myx_{t-1},\myz_t, \myw)$ }
    %\multicolumn{3}{c}{\includegraphics[width=0.9\linewidth]{architecture_labels_v2.pdf}}
    \end{tabular}
    \caption{Probabilistic dependencies on generation of $\myx_t$ with different mask design, the yellow block indicates the elements that will be masked in attention computation.}
    \label{fig:mask}
\end{figure*}

%% file: tab_supp_act_class_h36m.tex
\setlength{\tabcolsep}{4pt}
\begin{table}[h]
\begin{center}
\caption{Groups of actions of Human3.6m dataset, for training the action classifier.}
\label{table:group_15_class}
\resizebox{0.6\textwidth}{!}{\setlength{\tabcolsep}{0.8mm}{
\begin{tabular}{c|l}
\toprule
group number & \makecell[c]{original classes} \\
\midrule
0 & \makecell[l]{Directions, Discussion, Greeting, \\Photo, Posing, Purchases, WalkDog, Waiting}  \\
1 & Eating, Phoning, Sitting, Smoking  \\
2 &  SittingDown \\
3 &  Walking  \\
4 &  WalkTogether  \\
\bottomrule
\end{tabular}}}
\end{center}
\end{table}
\setlength{\tabcolsep}{1.4pt}

\setlength{\tabcolsep}{4pt}
\begin{table}[h]
\begin{center}
\caption{Implicit evaluations by different classification models on Human3.6m dataset. Our method always performs better.}
\label{table:res_15_class}
\resizebox{0.55\textwidth}{!}{\setlength{\tabcolsep}{0.8mm}{
\begin{tabular}{l|cc|cc}
\toprule
& \multicolumn{2}{c|}{5 groups} & \multicolumn{2}{c}{15 classes} \\
& \makecell[c]{Acc (\%) $\uparrow$} & \makecell[c]{FID $\downarrow$} & \makecell[c]{Acc (\%) $\uparrow$} & \makecell[c]{FID $\downarrow$}  \\
\midrule
Real data & 85.5 & - & 48.1 &- \\
\midrule
DLow~\cite{yuan2020dlow} & 65.9 & \textbf{1.412} & 22.7 & 1.566\\
gsps~\cite{mao2021gsps} & 65.0 & 2.030 & 22.2 & 1.915 \\
\midrule
ours & \textbf{70.0} & 1.708 & \textbf{28.1 } & \textbf{1.466}\\
\bottomrule
\end{tabular}}}
\end{center}
\end{table}
\setlength{\tabcolsep}{1.4pt}